\documentclass[letterpaper, 10 pt, conference]{ieeeconf}  

\IEEEoverridecommandlockouts                              

\usepackage{graphics} 
\usepackage{epsfig} 
\usepackage{mathptmx} 
\usepackage{times} 
\usepackage{amsmath} 
\usepackage{amssymb}  
\usepackage{tikz}

\usepackage{mathtools}
\usepackage{epstopdf}
\usepackage{tikz}
\usepackage{algorithm}
\usepackage{algpseudocode}
\usepackage{amsfonts}
\usepackage{caption}
\usepackage{subcaption}
\usepackage{dblfloatfix}
\usepackage{fixltx2e}

\DeclareMathOperator{\Tr}{Tr}
\newcommand{\norm}[1]{\left\lVert#1\right\rVert}

\newenvironment{remark}[1][Remark]{\begin{trivlist} \item[\hskip \labelsep {\bfseries #1}]}{\end{trivlist}}
\renewcommand{\baselinestretch}{0.97}

\title{\LARGE \bf A Convex Model of Humanoid Momentum Dynamics for \\ Multi-Contact Motion Generation}

\author{Brahayam Ponton$^{1}$, Alexander Herzog$^{1}$, Stefan Schaal$^{1}$$^{2}$ and Ludovic Righetti$^{1}$
\thanks{This research was mainly supported by the Max-Planck-Society, the European Research Council (ERC) under the European Union Horizon 2020 research and innovation programme (grant agreement No 637935) and the Max-Planck ETH Center for Learning Systems.}
\thanks{$^{1}$Max Planck Institute for Intelligent Systems - Autonomous Motion Department, Tuebingen, Germany
        \textsf{firstname.lastname@tuebingen.mpg.de}}%
\thanks{$^{2}$University of Southern California - Computational Learning and Motor Control Lab, Los Angeles, USA}%
}

\begin{document}
\maketitle
\thispagestyle{empty}
\pagestyle{empty}
%
\begin{abstract}
Linear models for control and motion generation of humanoid robots have received significant attention in the past years, not only due to their well known theoretical guarantees, but also because of practical computational advantages. However, to tackle more challenging tasks and scenarios such as locomotion on uneven terrain, a more expressive model is required. In this paper, we are interested in contact interaction-centered motion optimization based on the momentum dynamics model. This model is non-linear and non-convex; however, we find a relaxation of the problem that allows us to formulate it as a single convex quadratically-constrained quadratic program (QCQP) that can be very efficiently optimized. Furthermore, experimental results suggest that this relaxation is tight and therefore useful for multi-contact planning. This convex model is then coupled to the optimization of end-effector contacts location using a mixed integer program, which can be solved in realtime. This becomes relevant e.g. to recover from external pushes, where a predefined stepping plan is likely to fail and an online adaptation of the contact location is needed. The performance of our algorithm is demonstrated in several multi-contact scenarios for a humanoid robot.
\end{abstract}
%
\section{INTRODUCTION} \label{sec:introduction}
One of the major challenges that a humanoid robot faces when walking or running is keeping its balance. The difficulty arises from the fact that to move its Center of Mass (CoM) in a direction other than that of gravity, it needs to generate external forces by dynamically interacting with the environment through the creation of intermittent physical contacts. However, these contact forces, that allow to generate and control locomotion, are limited by the mechanical laws of unilateral contact \cite{NonsmoothMechanics}: feet can only push and not pull on the ground. This means that arbitrary motions are not possible, and therefore being able to continuously answer questions such as: where to place the feet, how hard to push, or in which direction to move the body? are important for generating safe and stable motions, even more in the case of uneven terrain or strong perturbations.

On one hand, simplified models (usually based on researchers' intuition about the simplified dynamics of humanoid robots, such as the linear inverted pendulum model (LIPM) \cite{conf/icra/KajitaKKFHYH03, DBLP:journals/tsmc/SardainB04}) provide an answer to these questions by formulating an optimal control problem, that exploits the linearity of the model and can be repeatedly solved in a receding horizon fashion as a quadratic program \cite{conf/icra/DimitrovPW11, conf/humanoids/Wieber06}. This model predictive control scheme endows robustness to the control strategy by bringing viability as a side effect of the optimization \cite{DBLP:conf/iros/Wieber08} and has made these approaches very successful in controlling locomotion of bipedal robots \cite{journals/trob/EnglsbergerOA15, DBLP:conf/humanoids/SherikovDW14}. However, simplified models have limitations, because of its assumptions such as co-planar footsteps, constant CoM height, zero angular momentum, among others, which might be undesirable in more dynamic maneuvers.

On the other hand, the benefits of using a full dynamics model have been demonstrated with the optimization of more complex behaviors as in \cite{DBLP:conf/wafr/PosaT12, DBLP:conf/iros/TassaET12}. While these approaches generate sophisticated whole body behaviors by making use of full rigid body dynamics models and simultaneous optimization of contact forces and robot motion, they are computationally expensive. Besides, in these formulations, the optimization landscape is very high-dimensional, prone to local-minima, non-convex and discontinuous due to contacts, characteristics that make the problem very hard to optimize.

Between these two extremes, there is a full range of model choices. An interesting approach is \cite{Herzog-2016b, DBLP:conf/humanoids/DaiVT14}, where a full kinematics model and the centroidal momentum dynamics model are combined. In this approach, the momentum generated kinematically and dynamically match, ensuring equivalence to the full dynamics model, provided that there exists enough torque authority. \cite{Herzog-2016b} further shows that only the momentum equations are necessary to reason about dynamics, which allows to solve the problem iteratively between kinematics and dynamics optimization. It is also shown that the optimal control problem using the centroidal momentum dynamics can be solved very efficiently (by exploiting the problem structure) as a sequence of convex QCQPs. \cite{JustinMomentumOptimization, AlexAuroPaper} show experimentally the utility of controlling momentum to stabilize and generate dynamic motion in a humanoid robot.

Other interesting line of research is the use of mixed-integer programs to optimize not only continuous postural adjustments and contact forces, but also discrete changes in the contact state. \cite{ibanez-IROS2014} shows the benefits of simultaneous adaptation of gait pattern and posture in a humanoid walking on flat ground. This work considers a simplified dynamics model (LIPM) to find footsteps that help the robot to maintain stability and locomote. \cite{DBLP:conf/humanoids/DeitsT14} does not consider a dynamics model, however it presents a method for footsteps planning on uneven terrain with obstacles using a mixed-integer quadratically-constrained quadratic program (MIQCQP). Integer decision variables are used to select out of a set of safe regions, the region over which to step. Another unique feature is the piecewise affine approximation of rotation, which keeps the form of the problem as a MIQCQP and makes it efficiently solvable to its global minimum.

In this paper, we are interested in planning: CoM motion and momentum, contact interaction forces, and a short sequence of contacts consistent with the desired dynamic motion (Fig. \ref{fig_001}). Specifically our contributions are:
\begin{enumerate}
\item We use the result presented in \cite{Herzog-2016b} (namely, that the torque contribution of each end-effector to the angular momentum rate can be analytically decomposed into a convex and a concave part) to find a convex relaxation of the angular momentum dynamics. This allows us to formulate the problem as a single convex QCQP. Our approach significantly improves computational complexity and can solve the momentum optimization problem in realtime. Moreover, experimental results suggest that the relaxation is tight: solutions of the relaxed convex problem correspond to the global optimum of the original, non-convex, problem.
\item Using our convex model, we extend \cite{DBLP:conf/humanoids/DeitsT14} by including a dynamic model and hand contacts. It allows to plan together contact locations, contact forces and momentum dynamics. Therefore, the contact plan is not blind to the dynamic evolution of the robot. Furthermore, it is fast enough to be used online for a short preview sequence of contacts.
\end{enumerate}
The remainder of this paper is structured as follows. In Sec. \ref{sec:problem_formulation}, we present the problem formulation. Then, in Sec. \ref{sec:technical_approach}, we show how to obtain a convex model of the momentum dynamics and present the extension of the contact planner to incorporate a dynamics model. We show experimental results in Sec. \ref{sec:experimental_results} and conclude the paper in Sec. \ref{sec:conclusion}.
\section{PROBLEM FORMULATION} \label{sec:problem_formulation}
\noindent The dynamic model of a floating-base rigid body system is
\begin{equation*}
   \textbf{H}(\textbf{q})\ddot{\textbf{q}} + \textbf{C}(\textbf{q},\dot{\textbf{q}}) = \textbf{S}^{T} \tau_{j} + \textbf{J}_{\textrm{e}}^{T} \lambda \enspace ,
\end{equation*}

\noindent where $\textbf{q}=\begin{bmatrix} q_{j}^{T} & x^{T} \end{bmatrix}^{T}$ denotes the robot state. $x \in SE(3)$ is the position and orientation of the floating base frame of the robot with respect to the inertial frame and $q_{j} \in \mathbb{R}^{n_{j}}$ are joints positions. $\textbf{H}(\textbf{q}) \in \mathbb{R}^{n_{j}+6 \times n_{j}+6}$ is the inertia matrix; $\textbf{C}(\textbf{q},\dot{\textbf{q}}) \in \mathbb{R}^{n_{j}+6}$ the vector of Coriolis, centrifugal, gravity forces; $\mathbf{S}=\begin{bmatrix} I^{n_{j} \times n_{j}} & 0 \end{bmatrix} \in \mathbb{R}^{n_{j} \times n_{j}+6}$ the selection matrix and represents the under-actuation of the system. $\tau_{j} \in \mathbb{R}^{n_{j}}$ is the vector of joint torques; $\mathrm{J}_{\mathrm{e}}$ the Jacobian of the contact constraints and $\lambda = \begin{bmatrix} \cdots & \mathbf{f}_{\mathrm{e}} & \tau_{\mathrm{e}} & \cdots \end{bmatrix}$ the vector of generalized forces, composed of forces $\mathbf{f}_{\mathrm{e}}$ and torques $\tau_{\mathrm{e}}$ acting at contact $e$.

The equations of motion can be decomposed into an actuated ${a}$ and an un-actuated $u$ part as follows:
\begin{subequations}
\begin{align}
   \textbf{H}_{\mathrm{a}}(\textbf{q})\ddot{\textbf{q}} + \textbf{C}_{\mathrm{a}}(\textbf{q},\dot{\textbf{q}}) &= \tau_{j} + \textbf{J}_{\mathrm{e,a}}^{T} \lambda \label{eq_actuated_part}\\
   \textbf{H}_{\mathrm{u}}(\textbf{q})\ddot{\textbf{q}} + \textbf{C}_{\mathrm{u}}(\textbf{q},\dot{\textbf{q}}) &= \textbf{J}_{\mathrm{e,u}}^{T} \lambda \label{eq_newton_euler}
\end{align}
\end{subequations}

Equation \eqref{eq_newton_euler} can be interpreted as the Newton-Euler equations of the system. It expresses the change of momentum of a robot as a function of external forces. Under the assumption of enough torque authority, any combination of forces $\lambda$ and accelerations $\ddot{\textrm{q}}$ can be realized, as shown by the actuated part of the equations of motion \eqref{eq_actuated_part}, if they are consistent with the underactuated dynamics \cite{WieberNonholonomy,AlexAuroPaper}. This suggests a natural decomposition for planning dynamically consistent multi-contact motions for legged robots: the momentum equations are sufficient to ensure dynamic feasibility and Equation \eqref{eq_actuated_part} is only necessary to ensure kinematic feasibility and torque limits \cite{AlexHumanoidsPaper}. The centroidal dynamics equations, when expressed at the robot CoM, are:
\begin{figure}
	\centering
	\includegraphics[width=.5\textwidth]{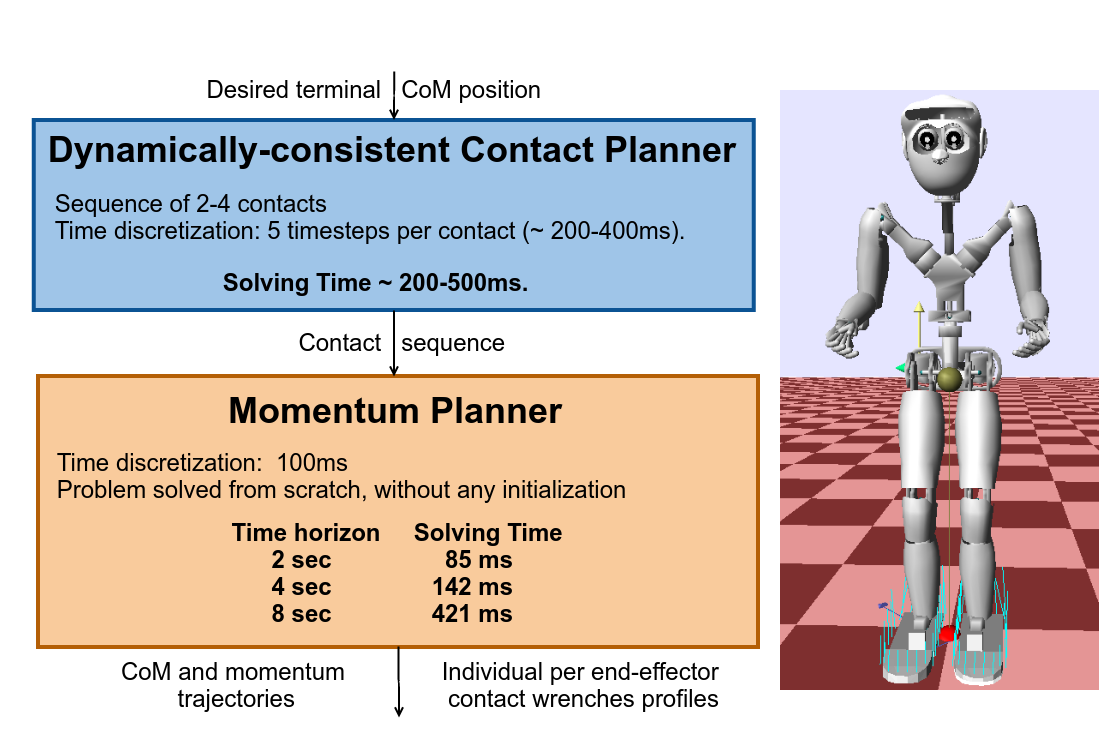}
	\caption{\small{Capabilities of the proposed algorithm.}} 
	\label{fig_001}
	\vspace{-0.5cm}
\end{figure}
\begin{equation}
\mathbf{\dot{h}} =
\begin{bmatrix}
    \mathbf{\dot{r}}  \\[0.0em]
    \mathbf{\dot{l}}  \\[0.0em]
    \mathbf{\dot{k}}  \\[0.0em]
\end{bmatrix} = 
\begin{bmatrix}
    \frac{1}{M} \mathbf{l}            \\[0.0em]
    M \mathbf{g} + \sum_{\text{e}} \mathbf{f}_{\text{e}}  \\[0.0em]
    \sum_{\text{e}} (\mathbf{p}_{\text{e}}+\mathbf{z}_{\text{e}}-\mathbf{r}) \times \mathbf{f}_{\text{e}} + \mathbf{\tau}_{\text{e}}
\end{bmatrix} \label{eqns_centroidal_momentum_dynamics}
\end{equation}

\noindent where $\mathbf{r}$, $\mathbf{l}$ and $\mathbf{k}$ denote the CoM position, linear and angular momenta, respectively. $M$ is the robot total mass and $\mathbf{g}$ the gravity vector. $\mathbf{z}_{\text{e}}$ is the center of pressure (CoP) position within the end-effector support region with respect to the point $\mathbf{p}_{\text{e}}$, denoting the position of the $e$ end-effector. Additionally, these dynamics are subject to physical constraints such as friction cones, CoPs within its support region in order to avoid tilting and torque limits.

In this paper, we concentrate on finding a convex formulation of the centroidal momentum dynamics \eqref{eqns_centroidal_momentum_dynamics}, to plan optimal dynamic motions and contact locations in realtime, which could then be realized by a low-level controller such as an inverse dynamics one \cite{AlexHumanoidsPaper}. Formally, we would like to find a solution that
\begin{equation}
\min_{\mathbf{p_{\mathrm{e}}}, \mathbf{z_{\mathrm{e}}}, \mathbf{f_{\mathrm{e}}}, \tau_{\mathrm{e}}} \quad \phi_{N}(\mathbf{h}_{N}) + \sum\limits_{t=1}^{N-1} \ell_{t}(\mathbf{h}, \mathbf{p_{\mathrm{e}}}, \mathbf{z_{\mathrm{e}}}, \mathbf{f_{\mathrm{e}}}, \tau_{\mathrm{e}}) \Delta_{t}
\label{eq_generic_cost}
\end{equation}

\noindent minimizes the sum of a terminal cost $\phi_{N}(\mathbf{h}_{N})$ and a running cost $\ell_{t}(\mathbf{h}, \mathbf{p_{\mathrm{e}}}, \mathbf{z_{\mathrm{e}}}, \mathbf{f_{\mathrm{e}}}, \tau_{\mathrm{e}})$, as will be defined later, over the available controls (namely contact locations $\mathbf{p_{\mathrm{e}}}$, CoP locations $\mathbf{z_{\mathrm{e}}}$, and contact wrenches  $\mathbf{f_{\mathrm{e}}}$,$\tau_{\mathrm{e}}$), under the discretized centroidal momentum dynamics:
\begin{equation}
\begin{bmatrix}
    \mathbf{r}_{\text{t}}        \\[0.0em]
    \mathbf{l}_{\text{t}}        \\[0.0em]
    \mathbf{k}_{\text{t}}        \\[0.0em]
    \mathbf{\dot{l}}_{\text{t}}  \\[0.0em]
    \mathbf{\dot{k}}_{\text{t}}  \\[0.0em]
 \end{bmatrix} = 
 \begin{bmatrix}
     \mathbf{r}_{\text{t}-1} + \frac{\Delta_{\text{t}}}{M} \mathbf{l}_{\text{t}}  \\[0.0em]         
     \mathbf{l}_{\text{t}-1} + \mathbf{\dot{l}}_{\text{t}} \Delta_{\text{t}}      \\[0.0em]
     \mathbf{k}_{\text{t}-1} + \mathbf{\dot{k}}_{\text{t}} \Delta_{\text{t}}      \\[0.0em]
     M \mathbf{g} + \sum_{\text{e}} \mathbf{f}_{\text{e,t}}                       \\[0.0em]
     \sum_{\text{e}} \mathbf{\kappa_{\text{e,t}}}                                 \\[0.0em]
  \end{bmatrix} \label{eq_linear_momentum}
\end{equation}
\noindent where the variable $\mathbf{\kappa_{\text{e,t}}}$ (end-effector contribution to angular momentum rate $\mathbf{\dot{k}}_{\text{t}}$) has been defined as
\begin{align}
\mathbf{\kappa_{\text{e,t}}} &= (\mathbf{p}_{\text{e}, \phi(\text{t})} + \mathbf{z}_{\text{e,t}} - \mathbf{r}_{\text{t}}) \times  \mathbf{f}_{\text{e,t}} + \mathbf{\tau}_{\text{e,t}}  \nonumber \\[0.1em]
&= {\ell}_{\text{e,t}} \times  \mathbf{f}_{\text{e,t}} + \mathbf{\tau}_{\text{e,t}}  \nonumber \\[0.1em] 
&=   \begin{bmatrix*}[r]
    0                               & -{\ell}^{\text{ z}}_{\text{e,t}} &  {\ell}^{\text{ y}}_{\text{e,t}}  \\[0.2em]
    {\ell}^{\text{ z}}_{\text{e,t}} &  0                               & -{\ell}^{\text{ x}}_{\text{e,t}}  \\[0.2em]
   -{\ell}^{\text{ y}}_{\text{e,t}} &  {\ell}^{\text{ x}}_{\text{e,t}} &  0                                \\[0.2em]
  \end{bmatrix*} 
  \begin{bmatrix}
    \mathbf{f}^{\text{ x}}_{\text{e,t}} \\[0.2em]
    \mathbf{f}^{\text{ y}}_{\text{e,t}} \\[0.2em]
    \mathbf{f}^{\text{ z}}_{\text{e,t}} \\[0.2em]
  \end{bmatrix} + \mathbf{\tau}_{\text{e,t}} \label{eq_torque_amomrate}
\end{align}

For notational simplicity we used the change of variable ${\ell}_{\text{e,t}} = (\mathbf{p}_{\text{e}, \phi(\text{t})} + \mathbf{z}_{\text{e,t}} - \mathbf{r}_{\text{t}})$. The variable $\phi(\text{t})$ has been introduced to denote that the end-effector position $\mathbf{p}_{\text{e}, \phi(\text{t})}$ remains fixed during a phase or sequence of predefined time steps. Physical constraints such as friction cone, CoP within region of support and torque limits are given by
\begin{subequations}
\begin{align}
&\norm{\prescript{_\mathrm{L}}{}{\mathbf{f}^{\text{ x}}_{\text{e,t}}}+\prescript{_\mathrm{L}}{}{\mathbf{f}^{\text{ y}}_{\text{e,t}}}} \le \mu_{\mathbf{f}} \prescript{_\mathrm{L}}{}{\mathbf{f}^{\text{ z}}_{\text{e,t}}}, \hspace{0.4cm} \prescript{_\mathrm{L}}{}{\mathbf{f}^{\text{ z}}_{\text{e,t}}} \ge 0, \label{eq_friction_cone} \\[0.3em]
&\prescript{_\mathrm{L}}{}{\mathbf{z}^{\text{ x}}_{\text{e,t}}} \in \left[ \mathbf{z}^{\text{ x}}_{\mathrm{min}}, \mathbf{z}^{\text{ x}}_{\mathrm{max}} \right], \hspace{1.05cm} \prescript{_\mathrm{L}}{}{\mathbf{z}^{\text{ y}}_{\text{e,t}}} \in \left[ \mathbf{z}^{\text{ y}}_{\mathrm{min}}, \mathbf{z}^{\text{ y}}_{\mathrm{max}} \right], \label{eq_conservative_cop}\\[0.3em]
&\norm{\prescript{_\mathrm{L}}{}{\tau^{\text{z}}_{\text{e,t}}}} \le \mu_{\tau} \prescript{_\mathrm{L}}{}{\mathbf{f}^{\text{ z}}_{\text{e,t}}} \label{eq_torque_cone}
\end{align}
\label{eq_friction_cop_torque}
\end{subequations}
\noindent where the left super-script $L$, denotes that the variables are expressed in local coordinate frames. \eqref{eq_friction_cone} expresses that forces belong to a friction cone with coefficient $\mu_{\mathbf{f}}$. \eqref{eq_conservative_cop} expresses that the CoP should be within a conservative region with respect to the real physical available region. \eqref{eq_torque_cone} constraints the torque to a cone with torsional coefficient $\mu_{\tau}$ \cite{DBLP:conf/icra/CaronPN15}. Torques $\prescript{_\mathrm{L}}{}{\tau^{\text{x}}_{\text{e,t}}}$ and $\prescript{_\mathrm{L}}{}{\tau^{\text{y}}_{\text{e,t}}}$ in local coordinate frames are zero. Local variables are mapped to the inertial frame through an appropriate rotation matrix $\mathbf{R}_{\mathrm{e,t}}$. We use right super-script on the rotation matrix to denote a particular set of its columns.
\begin{equation}
\begin{matrix*}[l]
\mathbf{f}_{\text{e,t}} = \mathbf{R}_{\mathrm{e,t}} \prescript{_\mathrm{L}}{}{\mathbf{f}_{\mathrm{e,t}}} \\
\tau_{\text{e,t}} = \mathbf{R}^{\mathrm{z}}_{\mathrm{e,t}} \prescript{_\mathrm{L}}{}{\tau^{\mathrm{z}}_{\mathrm{e,t}}} \\
\mathbf{z}_{\text{e,t}} = \mathbf{R}^{\mathrm{x,y}}_{\mathrm{e,t}} \prescript{_\mathrm{L}}{}{\mathbf{z}^{\mathrm{x,y}}_{\mathrm{e,t}}}
\end{matrix*}
\label{eq_map_local_to_world}
\end{equation}
Friction cones are usually approximated by a set of hyperplanes in pyramid shape. However, as we will write the dynamics as a convex QCQP, we do not require this approximation and keep the friction cone as a second order cone constraint on the contact forces. The only non-convexity in the described dynamics model comes from the variable $\mathbf{\kappa_{\text{e,t}}}$, which will be our focus in the next section.
\section{APPROACH} \label{sec:technical_approach}
\subsection{Reformulation of the dynamics}
Due to the practical and theoretical efficiency of solving a linear (LP) or a quadratic program, they have been exploited in many previous works, such as those where the angular momentum dynamics are neglected or the CoM height is kept constant (\cite{DBLP:conf/iros/AudrenVKEKY14, conf/icra/DimitrovPW11, journals/trob/EnglsbergerOA15, ibanez-IROS2014, conf/icra/KajitaKKFHYH03} to name a few). With few exceptions, the usual approach when the model complexity increases is to directly resort to general nonlinear solvers such as sequential quadratic programming or interior point methods, which might or might not find a solution for the non-convex problem \cite{JustinMomentumOptimization, DBLP:conf/wafr/PosaT12}, and also importantly, if they find one, the required time is not comparable to our formulation. In this paper, we find a convex approximation of the angular momentum dynamics, that allows to find solutions very efficiently. We will start by using the result presented in \cite{Herzog-2016b} that formulates the angular momentum dynamics as a difference of convex functions, and then we will analyze different alternatives to cope with the non-convex term.

\subsubsection*{Difference of Convex Functions Decomposition} The set of difference of convex functions $\mathcal{C^{\pm}}$ can be defined as:
\begin{align*}
\mathcal{C^{\pm}} = \bigg\{& \mathcal{C}^{+}(\mathbf{x}) - \mathcal{C}^{-}(\mathbf{x}) \;|\; \mathbf{x}\in\mathbb{R}^{n}, \big. \\
\big. &\mathcal{C}^{+}, \mathcal{C}^{-}: \mathbb{R}^{n} \rightarrow \mathbb{R}, \text{ are convex functions} \bigg\}
\end{align*}
\begin{figure}
	\centering
	\includegraphics[width=.30\textwidth]{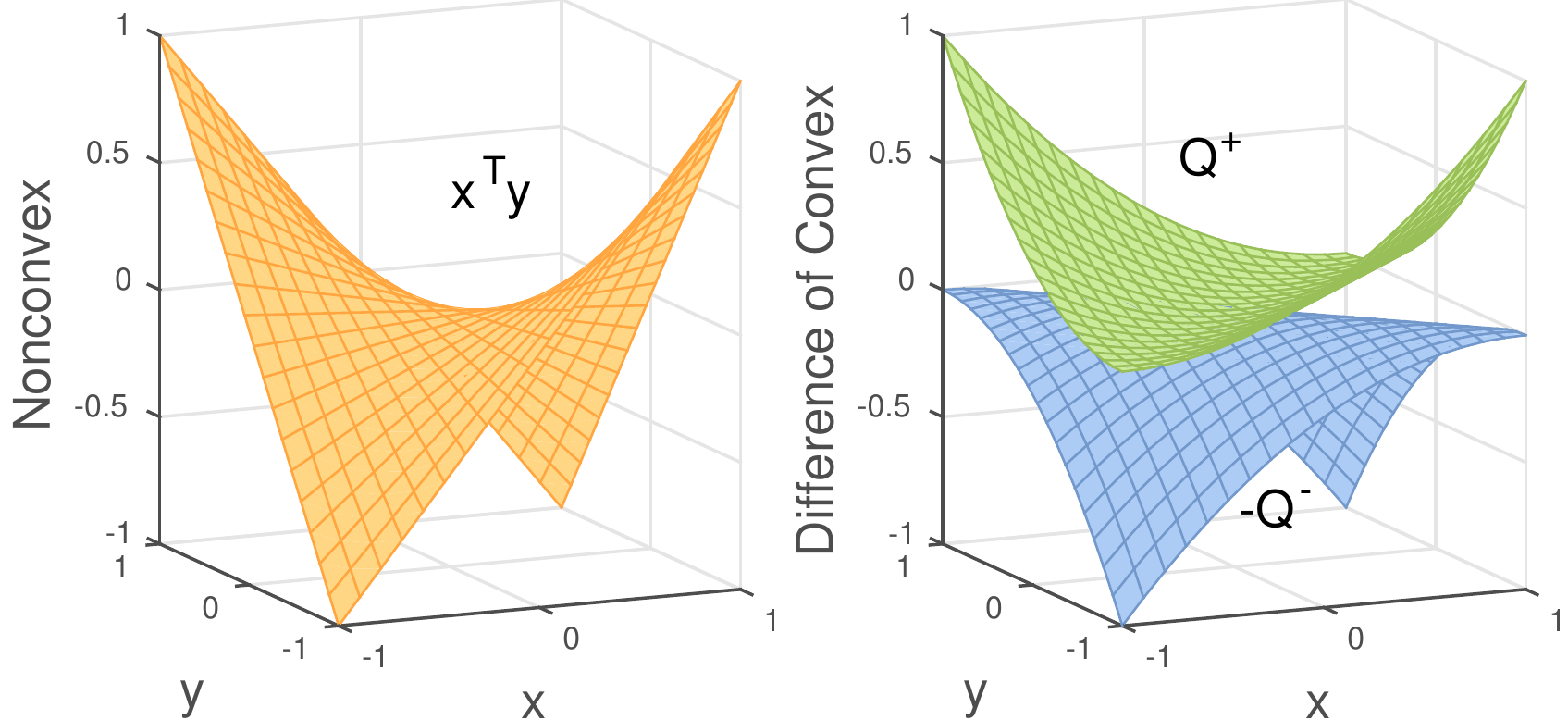}
	\vspace{-0.1cm}
	\caption{\small{Difference of convex functions decomposition.}} 
	\label{fig_002}
	\vspace{-0.6cm}
\end{figure}
This set is dense in the space of continuous functions \cite{LinearSystemsTheory}, which means that it can approximate with arbitrary accuracy any continuous function. Of particular interest are functions $\in \mathcal{C}^{\pm}$ that can be analytically decomposed, as is the case of scalar and cross products. Figure \ref{fig_002} shows an example decomposition of a scalar product $x^{T}y = Q^{+}-Q^{-}$, where
\begin{align*}
Q^{+} = \frac{1}{4} \norm{x+y}^{2} \text{, and } Q^{-} = \frac{1}{4} \norm{x-y}^{2}
\end{align*}

Using the example decomposition, it is easy to express each element of a cross product as an element of $\mathcal{C}^{\pm}$. First, each element is defined as a scalar product
\begin{align}
\mathbf{\ell \times \textbf{f}} &= 
   \begin{bmatrix}
     \underbrace{\begin{bmatrix} -{\ell}_{\text{z}} & \hspace{-0.1cm} {\ell}_{\text{y}} \end{bmatrix}}_{{a_{\text{cvx}}^{T}}^{'}}
     \overbrace{\begin{bmatrix} \mathbf{f}_{\text{y}} \\  \mathbf{f}_{\text{z}} \end{bmatrix}}^{{d_{\text{cvx}}}^{'}}, \hspace{-0.05cm}&
     \underbrace{\begin{bmatrix} {\ell}_{\text{z}} &  \hspace{-0.1cm} -{\ell}_{\text{x}} \end{bmatrix}}_{{b_{\text{cvx}}^{T}}^{'}}
     \overbrace{\begin{bmatrix} \mathbf{f}_{\text{x}} \\  \mathbf{f}_{\text{z}} \end{bmatrix}}^{{e_{\text{cvx}}}^{'}}, \hspace{-0.05cm}&
     \underbrace{\begin{bmatrix} -{\ell}_{\text{y}} & \hspace{-0.1cm} {\ell}_{\text{x}} \end{bmatrix}}_{{c_{\text{cvx}}^{T}}^{'}}
     \overbrace{\begin{bmatrix} \mathbf{f}_{\text{x}} \\  \mathbf{f}_{\text{y}} \end{bmatrix}}^{{f_{\text{cvx}}}^{'}} 
   \end{bmatrix}^{T}
   \label{eq_def_cvx_vars}
\end{align}
\noindent and then each scalar product is defined as an element of $\mathcal{C}^{\pm}$.
\begin{align}
\mathbf{\ell \times \textbf{f}} &= 
   \frac{\alpha}{4} 
   \begin{bmatrix}
     {\begin{Vmatrix} a_{\text{cvx}} +  d_{\text{cvx}} \end{Vmatrix}^{2} -
     \begin{Vmatrix} a_{\text{cvx}} - d_{\text{cvx}} \end{Vmatrix}^{2}} \\[0.15em]
     {\begin{Vmatrix} b_{\text{cvx}} +  e_{\text{cvx}} \end{Vmatrix}^{2} -
     \begin{Vmatrix} b_{\text{cvx}} - e_{\text{cvx}} \end{Vmatrix}^{2}} \\[0.15em]
     {\begin{Vmatrix} c_{\text{cvx}} +  f_{\text{cvx}} \end{Vmatrix}^{2} -
     \begin{Vmatrix} c_{\text{cvx}} - f_{\text{cvx}} \end{Vmatrix}^{2}} \\[0.15em]
   \end{bmatrix} \label{eq_length_cross_force}
\end{align}
\begin{remark}[Remark:]
When performing the decomposition, it is important not to forget the underlying physics of the system at hand. While the decomposition is mathematically correct and exploits the structure of the problem, the units of each of the terms of the decomposition do not match. Therefore, normalizing its variables is not only aesthetically pleasing, but also improves the conditioning of the problem, on which rate and region of quadratic convergence of Newton's method depend. Remember that the decomposition squares forces and lengths, which further increases their ratio.

In this problem, e.g. the variables $a_{\text{cvx}}^{'}$ and $d_{\text{cvx}}^{'}$ are defined in terms of lengths and forces, respectively. However, the variables $a_{\text{cvx}}$ and $d_{\text{cvx}}$ are defined in terms of normalized lengths and forces, respectively. The numerical constant $\alpha$ encodes the normalization of lengths by the nominal length of end-effectors (shoulder to hand distance for hands, and CoM to foot distance for feet) and forces by gravity.
\end{remark}

Now that the cross product has been decomposed, we will analyze alternatives to cope with its non-convexity $\mathcal{C}^{-}$. We will first introduce the standard method and then show how it can be improved for our purposes.

\subsubsection{Iterative Linearization}
This is the standard method to solve problems with constraints $\in \mathcal{C}^{\pm}$, and the one used in \cite{Herzog-2016b}. It reformulates the non-convex constraint as a convex one, by using a first-order Taylor expansion of the concave term evaluated at the current guess of the optimal vector. In the case of equality constraints, a slack variable is used to relax it to an inequality constraint, and also a penalty term over the slack variable is added to the cost. This procedure is performed iteratively until convergence to a solution. In a simple example, this procedure would look as follows:
\begin{align*}
\min_{x\in \mathbb{R}^{n}} \quad& f(x) \\
s.t. \quad& g(x) \le 0, \quad h(x) = 0, \quad x \in \Omega
\end{align*}
\noindent where $f(x)$ is the objective to be minimized, $g(x)$ is an inequality constraint, $h(x) \in \mathcal{C}^{\pm} = h^{+}(x) - h^{-}(x)$ is an equality constraint, and $\Omega$ is the feasible set, including upper and lower bounds on variables. Then, the problem to be solved iteratively would take the form:
\begin{align*}
\min_{x\in \mathbb{R}^{n}} \quad& f(x) + \mu \sum s_{i} \\
s.t. \quad& g(x) \le 0, \quad x \in \Omega, \quad s_{i} \ge 0  \\
     \quad& h^{+}(x) - \big( {h^{-}(x^{k}) + H^{-}_{k}(x-x^{k})}  \big)  \le s_{i}
\end{align*}
\noindent where $\mu$ is a penalty over the slack variables $s_{i}$ (if more than one constraint needed to be relaxed), and $H^{-}_{k} = \nabla_{x} \, h^{-}(x_{k})$ is the gradient of the concave term. To sum up, this method approximates at each iteration the concave term $\mathcal{C}^{-}$ by a locally valid hyperplane,  and neglects its second-order information (source of non-convexity). It is important to note that it does not neglect the entire concave part, because this would drastically change the constraint, instead it approximates it locally by a convex form. In the following, we present a new alternative procedure, that does not require to solve a sequence of convex programs, but a single one.
\subsubsection{Semidefinite Programming (SDP) Approach}
In this section, we present our first idea at trying to reformulate the non-convexity of the problem. The main idea is to upper bound the positive and negative definite components of the angular momentum dynamics. These upper bounds would then linearly define the angular momentum rate $\mathbf{\dot{k}}$ and would belong to a convex set. It is well known that, under some mild assumptions, convex problems can be solved efficiently in theory. However, convexity alone is not enough to guarantee the existence of efficient solution algorithms in practice. One type of convex set, namely the positive semidefinite cone $\mathbb{S}^{n}_{+}$, is amenable for efficient optimization and commonly used to compute lower bounds of integer problems.

In our setting, we will use SDP to compute every time step upper bounds for the positive and negative definite components of the angular momentum dynamics. In this way, the dynamics would be linear, and the upper bounds would be elements of the cone  $\mathbb{S}^{n}_{+}$. For simplicity of presentation, we denote $\mathbf{p}=a_{\text{cvx}}+d_{\text{cvx}}$ and $\mathbf{q}=a_{\text{cvx}}-d_{\text{cvx}}$ ($\mathbf{p}, \mathbf{q} \in \mathbb{R}^{2}$, see \eqref{eq_length_cross_force}) . Using this notation the $x$
 component of the torque contribution of an end-effector to the angular momentum rate dynamics \eqref{eq_torque_amomrate} becomes
\begin{align*}
   \kappa^{x} &= \frac{\alpha}{4} 
   \begin{bmatrix}
     \mathbf{p}^{T} \mathbf{p} - \mathbf{q}^{T} \mathbf{q}
   \end{bmatrix} + \tau^{x} \enspace,
\end{align*}
which is equivalent to the following formulation
\begin{align*}
   \kappa^{x} &= \frac{\alpha}{4} 
   \begin{bmatrix}
     \Tr{\begin{pmatrix} \mathbf{p} \mathbf{p}^{T} \end{pmatrix}} - \Tr{\begin{pmatrix} \mathbf{q} \mathbf{q}^{T} \end{pmatrix}}
   \end{bmatrix} + \tau^{x} \enspace,
\end{align*}
using the invariance of the trace under cyclic permutations:
\begin{align*}
\Tr\begin{pmatrix} \mathbf{p}\mathbf{p}^{T} \end{pmatrix} = \Tr\begin{pmatrix} \mathbf{p}^{T}\mathbf{p} \end{pmatrix} = \mathbf{p}^{T}\mathbf{p} \enspace .
\end{align*}
At this point, we introduce the variables $\mathbf{P}$, $\mathbf{Q} \in \mathbb{S}^{2}_{+}$ (they are $2\times2$ positive semidefinite matrices, which means that e.g. $x^{T} \mathbf{P} x \ge 0$) and perform the following change of variables
\begin{subequations}
\begin{align}
   &\kappa^{x} = \frac{\alpha}{4} 
   \begin{bmatrix}
     \Tr{\begin{pmatrix} \mathbf{P} \end{pmatrix}} - \Tr{\begin{pmatrix} \mathbf{Q} \end{pmatrix}}
   \end{bmatrix} + \tau^{x} \\
   &\mathrm{where:} \quad \mathbf{P} = \mathbf{p} \mathbf{p}^{T}, \quad \mathbf{Q} = \mathbf{q} \mathbf{q}^{T} \label{eq_change_of_var}
\end{align}
\end{subequations}
Now, we can point out exactly to the non-convexity and know its shape. Equation \eqref{eq_change_of_var} means that the feasible set is a cone surface (which is non-convex). The convex relaxation consist in using as feasible set the convex hull of this cone surface. This means that \eqref{eq_change_of_var} becomes:
\begin{align}
   &\mathbf{p} \mathbf{p}^{T} \preceq_{\mathbb{S}^{n}_{+}} \mathbf{P}, \quad \mathbf{q} \mathbf{q}^{T} \preceq_{\mathbb{S}^{n}_{+}} \mathbf{Q} \enspace, \label{eq_ineq_constraints_sdp}
\end{align}
In our specific problem, we can interpret the convex relaxation as follows: We have introduced new variables $\mathbf{P}$ and $\mathbf{Q}$, whose traces are upper bounds of $\mathbf{p}^{T} \mathbf{p}$ and $\mathbf{q}^{T} \mathbf{q}$ (that represent the positive and negative definite components of the contribution of each end-effector to the angular momentum rate). Then our work is to find conditions under which the gap (value difference between $\Tr{\begin{pmatrix} \mathbf{P} \end{pmatrix}} $ and $\Tr{\begin{pmatrix} \mathbf{Q} \end{pmatrix}} $ and the scalar products $\mathbf{p}^{T} \mathbf{p}$ and $\mathbf{q}^{T} \mathbf{q}$, respectively) is as small as possible. Making the gap between the upper bounds and the actual quantities zero means that the constraints in the non-convex problem are also satisfied exactly. In our problem, this would mean that the values of the torque contribution of each end-effector $\mathbf{\kappa_{\text{e,t}}}$ computed using the difference of upper bounds and using cross products of lengths and forces match. We defer this for the next section and will concentrate now on analyzing the current formulation.

In summary, we have relaxed the equality constraints to ($\mathbf{p} \mathbf{p}^{T} \preceq_{\mathbb{S}^{n}_{+}} \mathbf{P}$, $\mathbf{q} \mathbf{q}^{T} \preceq_{\mathbb{S}^{n}_{+}} \mathbf{Q}$), or equivalently $\mathbf{P}-\mathbf{p} \mathbf{p}^{T} \in \mathbb{S}^{n}_{+} \Leftrightarrow x^{T} (\mathbf{P}-\mathbf{p} \mathbf{p}^{T}) x \ge 0$ and $\mathbf{Q}-\mathbf{q} \mathbf{q}^{T} \in \mathbb{S}^{n}_{+} \Leftrightarrow x^{T} (\mathbf{Q}-\mathbf{q} \mathbf{q}^{T}) x \ge 0$. If we were to actually solve the problem using this formulation, the convex approximation of the end-effector torque contribution to the angular momentum rate dynamics using linear matrix inequalities (LMI) would be
\begin{subequations}
\begin{align}
   &\kappa^{x} = \frac{\alpha}{4} 
   \begin{bmatrix}
     \Tr{\begin{pmatrix} \mathbf{P} \end{pmatrix}} - \Tr{\begin{pmatrix} \mathbf{Q} \end{pmatrix}}
   \end{bmatrix} + \tau^{x} \\
   &\begin{bmatrix*}[l]
   \mathbf{P}     & \mathbf{p} \\
   \mathbf{p}^{T} & 1 \\
   \end{bmatrix*} \succeq 0, \quad
   \begin{bmatrix*}[l]
   \mathbf{Q}     & \mathbf{q} \\
   \mathbf{q}^{T} & 1 \\
   \end{bmatrix*} \succeq 0 \label{eq_lmi}
\end{align}
\end{subequations}

The transformation of inequalities \eqref{eq_ineq_constraints_sdp} to \eqref{eq_lmi} is performed using Schur's complement. What we can notice under the current formulation is that we have introduced new optimization variables $\mathbf{P}$ and $\mathbf{Q}$, which happen to be matrices, and it seems that we have only inflated the problem. This is true indeed; however, we have learned that
\begin{itemize}
\item A convex approximation of the angular momentum rate dynamics can be found using upper bounds of its positive and negative definite components.
\item We are approximating a scalar quantity (either $\mathbf{p}^{T}\mathbf{p}$ or $\mathbf{q}^{T}\mathbf{q}$), therefore using a matrix for it is unnecessary. A scalar quantity would suffice our purposes.
\item Positive semidefinite cones are too general, there are other cones, special cases of ${\mathbb{S}^{n}_{+}}$ for example, that we could use and more amenable to faster optimization.
\end{itemize}

\subsubsection{Quadratically-Constrained Quadratic Program Approach}
As mentioned in the last subsection, SDP includes e.g. as special cases LP (when the symmetric matrices involved are diagonal) or SOCP (when the symmetric matrices have an arrow form) \cite{ConvexOptimization}. In this subsection, we simplify the formulation of the previous subsection to a convex quadratic inequality constraint. By applying the linear trace operator on the previously defined inequalities \eqref{eq_ineq_constraints_sdp}, we would get
\begin{align*}
   \begin{matrix*}[l]
       \Tr(\mathbf{p} \mathbf{p}^{T}) \le \Tr(\mathbf{P})\enspace,  & \Tr(\mathbf{q} \mathbf{q}^{T}) \le \Tr(\mathbf{Q}) \\[0.3em]
       \mathbf{p}^{T} \mathbf{p} \le \Tr(\mathbf{P})\enspace,       & \mathbf{q}^{T} \mathbf{q} \le \Tr(\mathbf{Q}) \\[0.3em]
       \mathbf{p}^{T} \mathbf{p} \le \mathbf{\bar{p}}\enspace,      & \mathbf{q}^{T} \mathbf{q} \le \mathbf{\bar{q}}
   \end{matrix*}
\end{align*}

\noindent where we have introduced the scalar variables $\mathbf{\bar{p}}$, $\mathbf{\bar{q}} \in \mathbb{R}_{+}$ as the upper bounds of the positive and negative definite components of our dynamics constraint, which becomes
\begin{align*}
   &\kappa^{x} = \frac{\alpha}{4} 
   \begin{bmatrix}
     \mathbf{\bar{p}} - \mathbf{\bar{q}}
   \end{bmatrix} + \tau^{x} \\
   &\mathrm{where:} \quad \mathbf{p}^{T} \mathbf{p} \preceq_{\mathbb{R}_{+}} \mathbf{\bar{p}}, \quad \mathbf{q}^{T} \mathbf{q} \preceq_{\mathbb{R}_{+}} \mathbf{\bar{q}}
\end{align*}

Using this idea, the angular momentum dynamics of our problem can be reformulated as convex quadratic constraints:
\begin{align}
\mathbf{\kappa_{\text{e,t}}} &= {\ell}_{\text{e,t}} \times  \mathbf{f}_{\text{e,t}} + \mathbf{\tau}_{\text{e,t}} \nonumber \\ &= 
   \frac{\alpha}{4}
   \begin{bmatrix}
     \mathbf{u^{+}_{\text{x}}} - \mathbf{u^{-}_{\text{x}}} \\
     \mathbf{u^{+}_{\text{y}}} - \mathbf{u^{-}_{\text{y}}} \\
     \mathbf{u^{+}_{\text{z}}} - \mathbf{u^{-}_{\text{z}}} \\
   \end{bmatrix}_{\text{e,t}} + \mathbf{\tau}_{\text{e,t}} = \frac{\alpha}{4} \left[ \mathbf{U}^{+}_{\text{e,t}}-\mathbf{U}^{-}_{\text{e,t}} \right] + \mathbf{\tau}_{\text{e,t}}
   \label{eq_cvx_bnds}
\end{align}
where:
\begin{align}
  \begin{matrix}
    \begin{Vmatrix} a_{\text{cvx}} + d_{\text{cvx}} \end{Vmatrix}^{2}_{\text{e,t}} \preceq_{\mathbb{R}_{+}} \mathbf{u^{+}_{\text{x}}}_{\text{e,t}}, \qquad  
    \begin{Vmatrix} a_{\text{cvx}} - d_{\text{cvx}} \end{Vmatrix}^{2}_{\text{e,t}} \preceq_{\mathbb{R}_{+}} \mathbf{u^{-}_{\text{x}}}_{\text{e,t}}, \\[0.3em]
    \begin{Vmatrix} b_{\text{cvx}} + e_{\text{cvx}} \end{Vmatrix}^{2}_{\text{e,t}} \preceq_{\mathbb{R}_{+}} \mathbf{u^{+}_{\text{y}}}_{\text{e,t}}, \qquad
    \begin{Vmatrix} b_{\text{cvx}} - e_{\text{cvx}} \end{Vmatrix}^{2}_{\text{e,t}} \preceq_{\mathbb{R}_{+}} \mathbf{u^{-}_{\text{y}}}_{\text{e,t}}, \\[0.3em]
    \begin{Vmatrix} c_{\text{cvx}} + f_{\text{cvx}} \end{Vmatrix}^{2}_{\text{e,t}} \preceq_{\mathbb{R}_{+}} \mathbf{u^{+}_{\text{z}}}_{\text{e,t}}, \qquad
    \begin{Vmatrix} c_{\text{cvx}} - f_{\text{cvx}} \end{Vmatrix}^{2}_{\text{e,t}} \preceq_{\mathbb{R}_{+}} \mathbf{u^{-}_{\text{z}}}_{\text{e,t}}.
  \end{matrix} \label{eq_quadratic_constraints}
\end{align}
where $\mathbf{U^{+}_{\text{e,t}}}$ and $\mathbf{U^{-}_{\text{e,t}}}$ are upper bounds for each end-effector $e$ and time-step $t$ of the positive and negative definite components of the end-effector torque contribution $\mathbf{\kappa_{\text{e,t}}}$ to the angular momentum rate dynamics $\mathbf{\dot{k}}_{t}$. Notice that, in this formulation, differently from the iterative linearization, the concave part is not limited to take values within a hyperplane, but on the exact quadratic function, which is then upper bounded using a convex quadratic inequality constraint, instead of a LMI as in the SDP case.
\subsection{Cost function and summary} \label{subsec:cost_function}
In this subsection, we define the cost function to be optimized and summarize the optimization problem. The running cost $\ell^{d}_{t}$ of the dynamics optimization is given by:
\begin{align}
\ell^{d}_{t} = &\norm{\mathbf{l}_{t}}^{2}_{\mathcal{Q}_{\mathbf{l}}} + \sum_{\mathrm{e}} \norm{\mathbf{U}^{+}_{\text{t,e}}}^{2}_{\mathcal{Q}_{\mathbf{k}}} + \norm{\mathbf{U}^{-}_{\text{t,e}}}^{2}_{\mathcal{Q}_{\mathbf{k}}} + \nonumber \\
&\sum_{\mathrm{e}} \norm{\mathbf{f}_{\mathrm{e,t}}}_{\mathcal{Q}_{\mathbf{f}_{\mathrm{e}}}}+\norm{\prescript{_\mathrm{L}}{}{\tau^{\text{z}}_{\text{e,t}}}}_{\mathcal{Q}_{\tau_{\mathrm{e}}}} + \norm{\prescript{_\mathrm{L}}{}{\mathbf{z}^{\text{x,y}}_{\text{e,t}}}}_{\mathcal{Q}_{\mathbf{z}_{\mathrm{e}}}}
\label{eq_dynamics_cost}
\end{align}
were a cost of the form $\norm{x}^{2}_{\mathcal{Q}_{x}}$ represents a quadratic cost $x^{T} \mathcal{Q}_{x} x$, with $\mathcal{Q}_{x}$ a positive semi-definite matrix. This cost penalizes high forces $\mathbf{f}_{\mathrm{e,t}}$ and torques $\prescript{_\mathrm{L}}{}{\tau^{\text{z}}_{\text{e,t}}}$, deviations of the CoP $\prescript{_\mathrm{L}}{}{\mathbf{z}^{\text{x,y}}_{\text{e,t}}}$ from the end-effector position $\mathbf{p}_{\mathrm{e},\phi(t)}$ (such that it stays close to the center of the support region). This cost also includes a capturability penalty by penalizing a derivative of the CoM position \cite{DBLP:conf/iros/Wieber08}, namely, the linear momentum $\mathbf{l}_{t}$.

The terminal cost $\phi_{N}(\mathbf{h}_{N})$ is usually defined as the point where we would like to be at the end of the time horizon or a velocity we would like to track.
\begin{remark}[Remark]
In \eqref{eq_dynamics_cost}, we did not include directly a cost over the angular momentum $\mathbf{k}_{t}$ (e.g. $\norm{\mathbf{k}_{t}}^{2}_{\mathcal{Q}_{\mathbf{k}}}$), because this variable is defined in terms of its rate $\mathbf{\dot{k}}_{t}$; which in turn is defined as a sum of per end-effector upper bounds of positive $\mathbf{U}^{+}_{\text{e,t}}$ and negative $\mathbf{U}^{-}_{\text{e,t}}$ definite components \eqref{eq_cvx_bnds}; therefore, if we were to penalize either $\mathbf{k}_{t}$ or $\mathbf{\dot{k}}_{t}$ directly (e.g. $\norm{\mathbf{\dot{k}}_{t}}^{2}_{\mathcal{Q}_{\mathbf{\dot{k}}}}$), they could be made trivially zero by the upper bounds ($\mathbf{U}^{+}_{\text{e,t}}$, $\mathbf{U}^{-}_{\text{e,t}}$) taking any value higher than the scalar product they upper bound. In this case, there would be a gap between upper bounds and the actual positive and negative definite components defined by the scalar products \eqref{eq_quadratic_constraints}. Therefore, in practice we penalize angular momentum indirectly by adding a cost over the upper bounds ($\mathbf{U}^{+}_{\text{e,t}}$ and $\mathbf{U}^{-}_{\text{e,t}}$), which is a penalization over the contribution to angular momentum rate per end-effector, separately for its positive and negative definite part. In this case, we have empirically found that the gap is, up to numerical precision, zero. Intuitively, you can think of the upper bounds as free variables that can take any value higher than the quantity they upper bound. However, as they are penalized, they will not be higher than needed by the angular momentum and will actually be on the cone surface, making the approximation tight.
\end{remark}
To ensure physical consistency at execution time, we also include the constraint
\begin{align}
\norm{\mathbf{p}_{\mathrm{e},\phi(t)} - \mathbf{r}_{\mathrm{t}}} \le \ell_{\mathrm{e}}^{\mathrm{max}} \label{eq_eef_len} \enspace ,
\end{align}
\noindent that constraints the distance from the CoM to the end-effector position by the maximum end-effector length $\ell_{\mathrm{e}}^{\mathrm{max}}$ (previously called nominal length, used for normalization). This constraint holds as is for feet, but in the next section, we will show how it is adapted for hands.

The following convex QCQP program summarizes the optimization problem assuming a fixed set of contacts:
\begin{equation}
\begin{aligned}
& \text{minimize}
& & \phi_{N}(\mathbf{h}_{N}) + \sum_{t} \ell^{d}_{t} \Delta_{\text{t}} \\
& \text{subject to}
& & \eqref{eq_linear_momentum}, \eqref{eq_friction_cop_torque}, \eqref{eq_map_local_to_world}, \eqref{eq_def_cvx_vars}, \eqref{eq_cvx_bnds}, \eqref{eq_quadratic_constraints}, \eqref{eq_eef_len} \quad \forall \text{ t,e} \enspace.  \\
\label{dynamics_problem}
\end{aligned}
\vspace{-0.5cm}
\end{equation}
The optimization variables are: CoM $\mathbf{r}_{\text{t}}$, linear momentum and its rate $\mathbf{l}_{\text{t}}$, $\mathbf{\dot{l}}_{\text{t}}$, angular momentum and its rate $\mathbf{k}_{\text{t}}$, $\mathbf{\dot{k}}_{\text{t}}$, wrenches in local and inertial coordinate frames $\mathbf{f_{\mathrm{\text{e,t}}}}$, $\tau_{\mathrm{\text{e,t}}}$, $\prescript{_\mathrm{L}}{}{\mathbf{f}_{\text{e,t}}}$, $\prescript{_\mathrm{L}}{}{\tau^{\text{z}}_{\text{e,t}}}$, CoP in local and inertial coordinate frames $\mathbf{z_{\mathrm{\text{e,t}}}}$,  $\prescript{_\mathrm{L}}{}{\mathbf{z}^{\text{x,y}}_{\text{e,t}}}$, upper bounds $\mathbf{U}^{-}_{\text{e,t}}$, $\mathbf{U}^{+}_{\text{e,t}}$, and per time-step and per end-effector auxiliary variables $a_{\text{cvx}}$, $b_{\text{cvx}}$, $c_{\text{cvx}}$, $d_{\text{cvx}}$, $e_{\text{cvx}}$, $f_{\text{cvx}}$. From \eqref{eq_def_cvx_vars}, we only use the definition of auxiliary variables $a_{\text{cvx}}$, $b_{\text{cvx}}$, and $c_{\text{cvx}}$ in terms of lengths $\ell_{\text{e,t}}$, and $d_{\text{cvx}}$, $e_{\text{cvx}}$, and $f_{\text{cvx}}$ in terms of forces in world coordinate frame. They are appropriately normalized, therefore, the primed variables can be replaced by non-primed ones (e.g $a_{\text{cvx}}^{'} \rightarrow a_{\text{cvx}}$). Equation \eqref{eq_dynamics_cost} defines the running cost $\ell^{d}_{t}$.

\subsection{Contacts planning}
Up to this point, we have described an optimization problem able to efficiently plan CoM motion and interaction forces of the robot with the environment  given a plan of non-coplanar contacts. This optimization problem, formally described as a convex QCQP, can be easily embedded in the footstep planner algorithm \cite{DBLP:conf/humanoids/DeitsT14}. The main adaptation would be in the definition of the cost, where the goal would no longer be finding a possibly large sequence of footsteps such that a desired final feet configuration (position and orientation) is achieved, but instead finding a short sequence of contacts that support the achievement of a dynamic motion. The optimization problem would be to minimize the cost given by $\ell_{t}^{d}$ \eqref{eq_dynamics_cost} plus a regularization of the distance between footsteps under the constraints imposed by the dynamics model \eqref{dynamics_problem} and the contacts planner \cite{DBLP:conf/humanoids/DeitsT14}. Of course, a simplification of the contacts planner constraints is possible, given that its purpose differs from the original formulation. Exploiting the fact that, in this formulation the CoM of the robot is a decision variable, we make an extension of the algorithm, by including in the optimization, the search for contacts using hands. In the following, we briefly introduce algorithm \cite{DBLP:conf/humanoids/DeitsT14} and its extension.
\subsubsection{Original formulation}
\begin{figure}
	\centering
	\includegraphics[width=.26\textwidth]{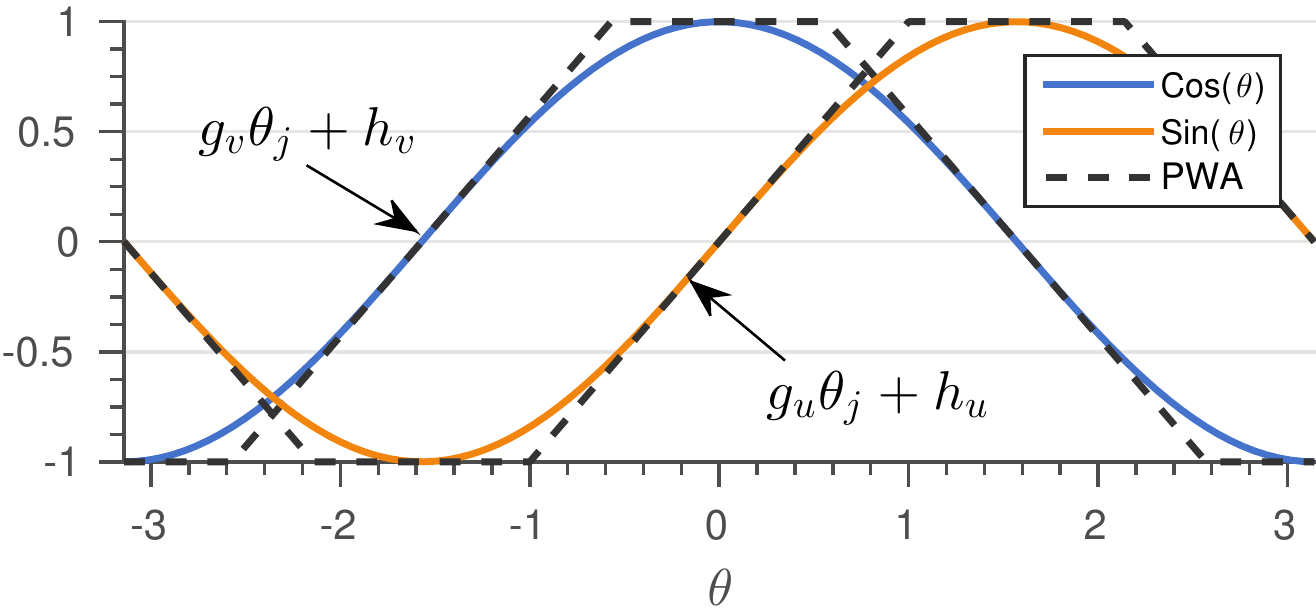}
    \vspace{-0.3cm}
	\caption{\small{Piecewise affine approximation of sine and cosine.}} 
	\label{fig_003}
	\vspace{-0.5cm}
\end{figure}

In \cite{DBLP:conf/humanoids/DeitsT14}, the terrain description consists of a set of convex, obstacle free regions $r \in \left\{ 1,R \right\}$, and the optimization considers a sequence of $j = \phi(t) \in \left\{ 1,n \right\}$ footsteps. For each footstep $j$, a piecewise affine (PWA) approximation of sine and cosine is used in order to handle footstep rotation (see Fig.\ref{fig_003}). 
\begin{align*}
  \begin{matrix*}[l]
    S_{u,j} \implies \begin{cases} \phi_{u} \le \theta_{j} \le \phi_{u+1} \\ s_{j} = g_{u} \theta_{j} + h_{u}  \end{cases} & u=1,...,U \\[0.3em]
    C_{v,j} \implies \begin{cases} \phi_{v} \le \theta_{j} \le \phi_{v+1} \\ c_{j} = g_{v} \theta_{j} + h_{v}  \end{cases} & v=1,...,V \\[0.3em]
  \end{matrix*}
\end{align*}
\noindent $u \in \left\{ 1,U \right\}$ and $v \in \left\{ 1,V \right\}$ are the number of linear functions used in the approximation. $S_{u,j}$ and $C_{v,j}$ are integer decision variables that define the active sine and cosine linear function for footstep $j$. The intersection of $\theta_{j} \in [\phi_{u}, \phi_{u+1}]$ and $\theta_{j} \in [\phi_{v}, \phi_{v+1}]$ is the region of validity of the approximation and, $s_{j} = g_{u} \theta_{j} + h_{u}$ and $c_{j} = g_{v} \theta_{j} + h_{v}$ are the linear approximations. Another set of integer decision variables $H_{r,j}$ defines the region $r \in \left\{ 1,R \right\}$, whose domain contains footstep $j$
\begin{align*}
  \begin{matrix*}[l]
    H_{r,j} \implies A_{r} \mathbf{p}_{j} \le b_{r} & r=1,...,R \enspace,
  \end{matrix*}
\end{align*}
\noindent where the inequality constraint $A_{r} \mathbf{p}_{j} \le b_{r}$, constraints the footstep position $\mathbf{p}_{j}$ to not only belong to the epigraph defined by region $r$, but also to be lying on it, because all footsteps in the plan are active. Additionally,  integrality constraints are imposed over the integer decision variables, which renders the optimization efficient:
\begin{align*}
& \begin{matrix*}[l]
    \sum_{r} \; H_{r,j} = \sum_{u} \; S_{u,j} = \sum_{v} \; C_{v,j} = 1 &\\[0.5em]
    H_{r,j}, \;\; S_{u,j}, \;\; C_{v,j} \;\; \in \;\; \{0,1\} & \\[0.1em]
  \end{matrix*}
\end{align*}
This constraints each footstep to belong only to one region and to use only one linear model to approximate the rotation. Finally, another remarkable feature of \cite{DBLP:conf/humanoids/DeitsT14}, is the expression of reachability constraints between footstep positions as the intersection of SOCP constraints:
\begin{align*}
& \norm{  \begin{bmatrix} \mathbf{p}^{x}_{j} \\ \mathbf{p}^{y}_{j} \end{bmatrix} - \left( \begin{bmatrix} \mathbf{p}^{x}_{j-1} \\ \mathbf{p}^{y}_{j-1} \end{bmatrix} + \begin{bmatrix} c_{j} & -s_{j} \\ s_{j} & c_{j} \end{bmatrix} r_{i}    \right)   } \le d_{i}, \;\; i=1,2 \enspace ,
\end{align*}
\noindent where $r_{i} \in \mathbb{R}^2$ for $i=1,2$ are distances in opposite direction from the last end-effector position $\mathbf{p}_{j-1}$ and can also be rotated by $\theta$. They define new points, whose distance to the next footstep position $\mathbf{p}_{j}$ cannot exceed $d_{i}$.

\subsubsection{Extension to hand end-effectors}

In the following, we present the extension of the algorithm, consisting in defining reachability and safe region constraints for hands, and including an additional set of integer decision variables that define the activation of hand end-effectors.

Hands, in the same way as feet, are also constrained to belong to a unique safe region $r$ at each phase $j$; however, as they cannot always be in contact, for them belonging to a safe region has the meaning of being in the epigraph of the region. Let's denote by $\mathbf{n}_{r}$ the outward-pointing normal of the surface of a region, by $\mathbf{s}_{r}$ any point on the surface, and by $\mathbf{A}^{h}_{r}$, $\mathbf{b}^{h}_{r}$ the hyperplanes that define the borders of the region. For a hand the safe region constraint is defined as:
\begin{align*}
& A_{r} \mathbf{p}_{j} \le b_{r} \\
\text{where: }   
  &A_{r} = \begin{bmatrix*}[r]
    \mathbf{A}^{h}_{r} \\
    -\mathbf{n}_{r}
  \end{bmatrix*}
\text{, and }   
b_{r} = \begin{bmatrix}
    \mathbf{b}^{h}_{r} \\
    -\mathbf{n}_{r}^{T}\mathbf{s}_{r}
  \end{bmatrix} \enspace,
\end{align*}
\noindent while for a foot, it also contains the constraint $\mathbf{n}_{r}^{T}\mathbf{p}_{j} \le \mathbf{n}_{r}^{T}\mathbf{s}_{r}$, that constraints the contact to be lying on the surface.

To define reachability constraints for hands, we introduce an additional vector $\mathbf{r}_{tr}$, defined as the difference between the position of the upper torso and the CoM. This vector has constant length, but encodes the current orientation of the torso. This vector is updated every time we re-optimize the contact plan, but is kept constant during the optimization. With this in mind, the reachability constraint becomes:
\begin{align*}
& \norm{  \mathbf{p}_{j} - \left( \mathbf{r}_{j} + \mathbf{r}_{tr} + \begin{bmatrix} c_{j} & -s_{j} & 0 \\ s_{j} & c_{j} & 0 \\ 0 & 0 & 1\end{bmatrix} \mathbf{r}_{sh}    \right)   } \le \ell^{\text{ max}}_{arm} \enspace.
\end{align*}
%
%
\begin{figure}
	\centering
	\centering
    \includegraphics[trim=100 100 100 0, clip, width=.08\textwidth]{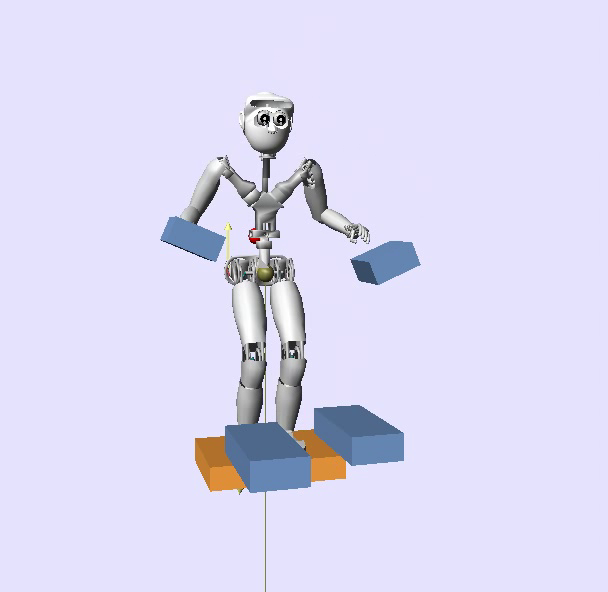}
	\includegraphics[trim=100 100 100 0, clip, width=.08\textwidth]{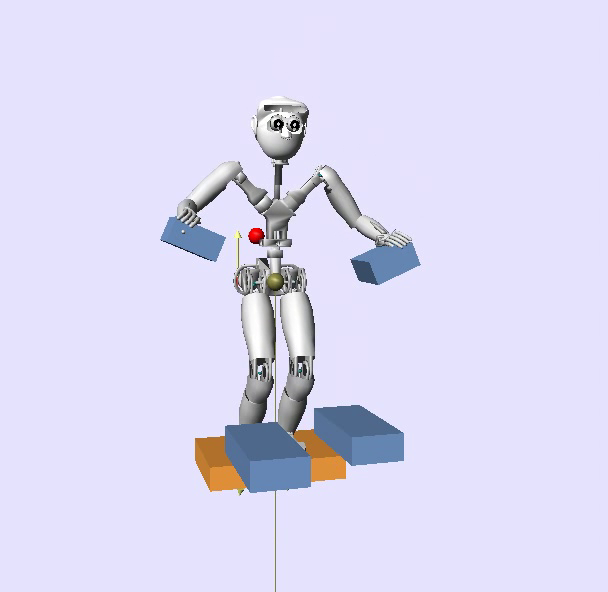}
	\includegraphics[trim=100 100 100 0, clip, width=.08\textwidth]{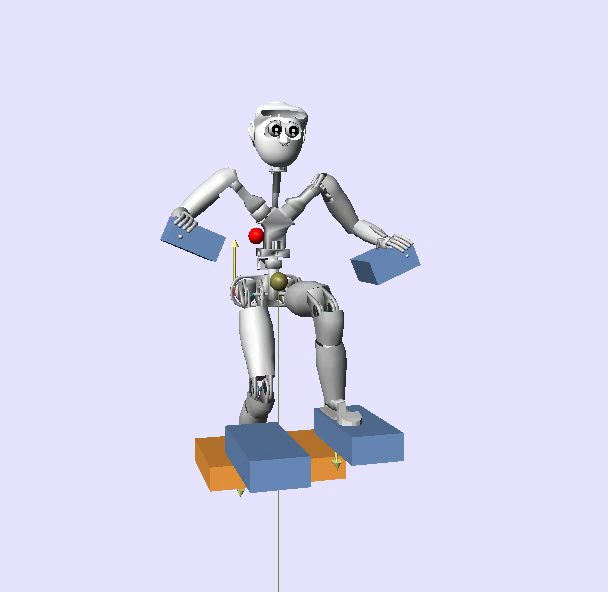}
	\includegraphics[trim=100 100 100 0, clip, width=.08\textwidth]{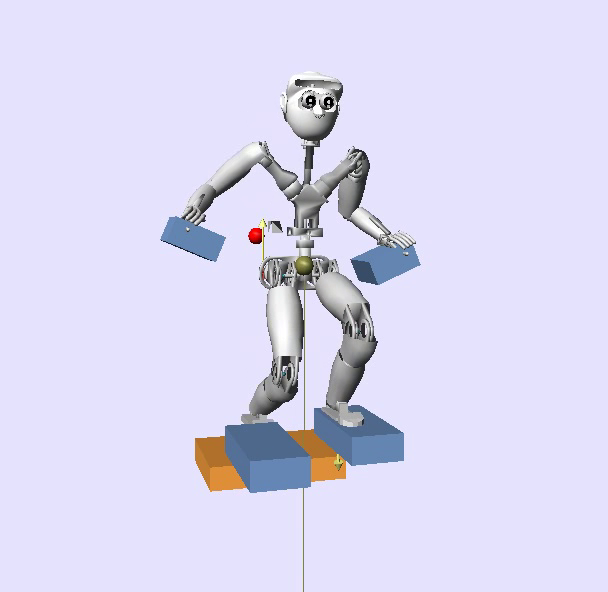}
	\includegraphics[trim=100 100 100 0, clip, width=.08\textwidth]{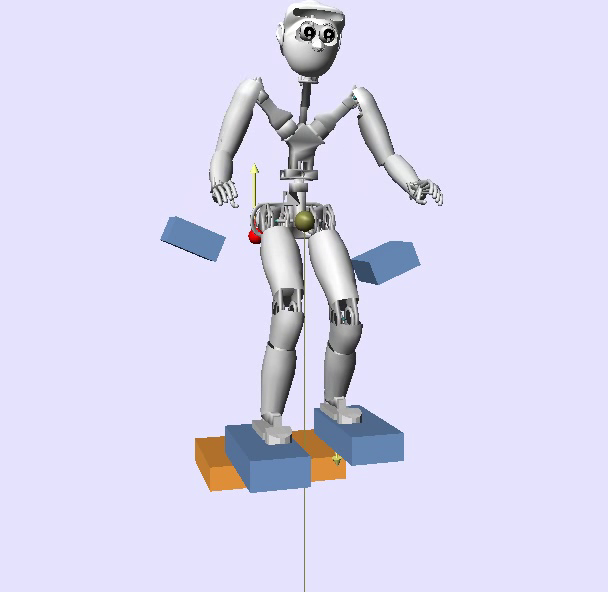}
    \caption{Climbing an uneven terrain using hands}
    \label{fig:task2}
    \vspace{-0.7cm}
\end{figure}

%
$\mathbf{p}_{j}$ is the position of a hand, $\mathbf{r}_{sh}$ is a constant vector pointing from the upper torso to the shoulder of the corresponding hand, and $\ell^{\text{ max}}_{arm}$ is the maximum distance between shoulder and hand. Therefore, the term within the norm is the relative position of a hand with respect to its shoulder. In the objective function, we include regularization of this term from a default relative position. A set of integer decision variables $J \in \mathbb{R}^{2 \times n}$ per hand is also used to define contact activations.
\begin{align*}
  \begin{matrix*}[l]
    J_{1,j} \implies \begin{cases} \mathbf{p}_{\mathrm{e,j}} = \mathbf{p}_{\mathrm{e,j-1}}  \\
    \mathbf{n}_{r}^{T}\mathbf{p}_{j} \le \mathbf{n}_{r}^{T}\mathbf{s}_{r}
    \end{cases}
    J_{2,j} \implies \begin{cases} \prescript{_\mathrm{L}}{}{\mathbf{f}_{\text{e,t}}} = 0 \\ \mathbf{p}_{\mathrm{e,j}} \neq \mathbf{p}_{\mathrm{e,j-1}}
    \end{cases}
  \end{matrix*}
\end{align*}
valid $\forall t \in \phi(t)=j$. The columns of $J$ are constrained to sum one: $\sum_{i}J_{i,j}=1, \forall j$. Either the end-effector is in contact or not. The problem lies on the fact that, it is not possible to determine how many hand activations are necessary during a task. Therefore, the sum of the first row of $J$ has to be left free $\sum_{j}J_{1,j}\le \text{n}_{\text{max}}$. This is the fact that limits the planner to only a short sequence of contacts, usually $n=n_{max}=(\text{3 or 4})$, in order to obtain results in a reasonable time.

Finally, the decision variables on safe regions, also affect the mapping of quantities from local to world frame, by the selection of the appropriate rotation matrix in \eqref{eq_map_local_to_world}.
\begin{align*}
  \begin{matrix*}[l]
    H_{r,j} \implies \begin{cases} \mathbf{f}_{\text{e,t}} = \mathbf{R}_{\mathrm{e,t}\in\phi(t)} \prescript{_\mathrm{L}}{}{\mathbf{f}_{\mathrm{e,t}}} \\
    \mathbf{z}_{\text{e,t}} = \mathbf{R}^{\mathrm{x,y}}_{\mathrm{e,t}\in\phi(t)} \prescript{_\mathrm{L}}{}{\mathbf{z}^{\mathrm{x,y}}_{\mathrm{e,t}}} \\
    \tau_{\text{e,t}} = \mathbf{R}^{\mathrm{z}}_{\mathrm{e,t}\in\phi(t)} \prescript{_\mathrm{L}}{}{\tau^{\mathrm{z}}_{\mathrm{e,t}}}
    \end{cases}
  \end{matrix*}
\end{align*}

The mapping of discrete quantities $j$ to continuous time is fixed by predefining the mapping $j=\phi(t)$, that defines the timing of each discrete phase in terms of the time $t$.
\addtolength{\textheight}{+0.5cm}  
%
%
\section{EXPERIMENTAL RESULTS} \label{sec:experimental_results}
In this section, we present experimental evaluations of our algorithm (Figure \ref{fig_001}) on several multi-contact scenarios 
for a simulated humanoid robot: climbing on an uneven terrain using only feet, walking on an uneven terrain, walking on the
same terrain with an external disturbance, climbing using hands and feet, traversing monkey bars and avoiding an obstacle using hands and feet. The results of the experiments are visible in the accompanying video \footnote{It can also be found under https://youtu.be/qLrftO0w5g4}.

All the tasks have a time duration of around 10 seconds. The contact planner first finds a sequence of dynamically consistent contact locations. We use 5 timesteps between each contact change (timestep duration between 200 and 400ms). In a second stage
the CoM, momentum and contact forces are optimized using the contact sequence. The time discretization in this case is finer
(100ms). The whole plan is computed without any initial guess (only the final desired CoM location is specified). For all the
tasks the complete plan is found at once, except for the push recovery task where the complete plan for the next two contacts
is computed using a receding horizon of 200ms.
Both optimization problems were solved using the parallel barrier algorithm implemented in Gurobi which
is particularly efficient for second order cone constraints and MIQCQP\footnote{We found the same solutions using Snopt and Ipopt but the required time was orders of magnitude larger.}.
We use inverse kinematics (tracking both momentum and end-effector positions) to visualize the whole-body motion of the robot
following the dynamic plan.

\begin{figure}
	\centering
	\includegraphics[width=.25\textwidth]{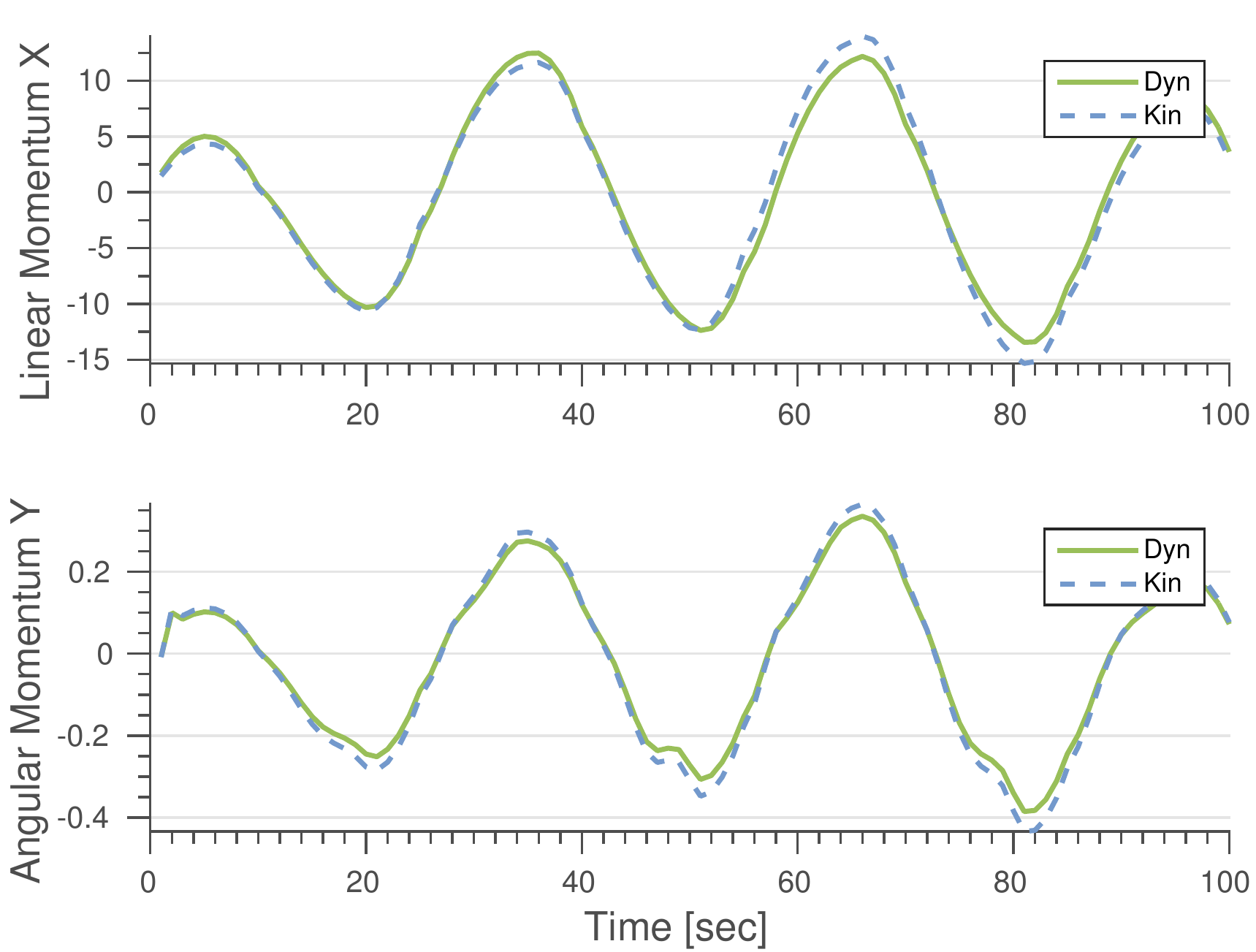}
    \vspace{-0.2cm}
	\caption{\small{Example momentum trajectories for a walking motion.n}} 
	\label{fig_005}
	\vspace{-0.7cm}
\end{figure}

Examples of the linear and momentum trajectories for a walking motion are shown in Fig. \ref{fig_005}. Green lines depict trajectories optimized using the dynamic formulation presented in this paper. Blue lines show the momentum trajectories optimized kinematically to track the desired dynamic trajectories (used for videos). As can be seen, in this figure and in the videos, momentum trajectories are non-trivial. In this particular case, we show the linear momentum in the direction responsible for moving the CoM laterally between footsteps, and the angular momentum in the forward direction of walking, responsible for the motion of arms.

\subsection{Solution Time and Computational Complexity}
The computational complexity $\mathcal{C}$ of solving a dense but convex QCQP using a primal-dual interior point method with an $m$-self-concordant function is polynomial and of the order $\mathcal{O}(m^{\frac{1}{2}}[m+n]n^{2})$ \cite{InteriorPointPTM}, where $m$ is the number of convex quadratic inequalities and $n$ the size of the optimization vector. However, in the case of a sparse problem (which is our case) we can expect to have a better computational complexity. 
Our formulation requires 9 quadratic inequalities per end-effector and time-step (6 to upper and lower bound the end-effector contribution to the angular momentum rate \eqref{eq_quadratic_constraints}, 1 to constraint the distance from the CoM \eqref{eq_eef_len}, 2 for friction and torque constraints \eqref{eq_friction_cone}, \eqref{eq_torque_cone}). Note that they are part of the optimization, only if end-effector $e$ is active at timestep $t$. 

All the tasks were optimized in around 1 sec (including both contact and momentum optimization). This is, to the best of our knowledge, much faster (at least one order of magnitude and much more in other cases) than other approaches to optimize motion for a humanoid robot under non-trivial conditions \cite{JustinMomentumOptimization, DBLP:conf/humanoids/DaiVT14, Herzog-2016b, journals/tog/MordatchTP12, DBLP:conf/wafr/PosaT12}.

For the contact planner (MIQCQP), as mentioned previously, we look ahead for 2 to 4 contacts and use a rough granularity for the dynamics. Depending on the number of terrain regions, the time required to find a contact plan varies between 200 and 500 milliseconds (See Fig. \ref{fig_001}). The quadratic inequality constraint \eqref{eq_eef_len} helped to quickly discard infeasible stepping regions.

\begin{figure}[H]
  \footnotesize
  \begin{minipage}[b]{0.50\linewidth}
    \centering%
      \begin{tabular}{||c c||} 
        \hline
        Timesteps & Time[ms] \\ [0.1ex] 
        \hline\hline  
        10 & 40 \\ 
        20 & 85 \\
        40 & 142 \\
        80 & 421 \\
        \hline
      \end{tabular}
    \par\vspace{2pt}
  \end{minipage}
  \begin{minipage}[b]{0.49\linewidth}
    \centering
    \includegraphics[width=0.6\linewidth]{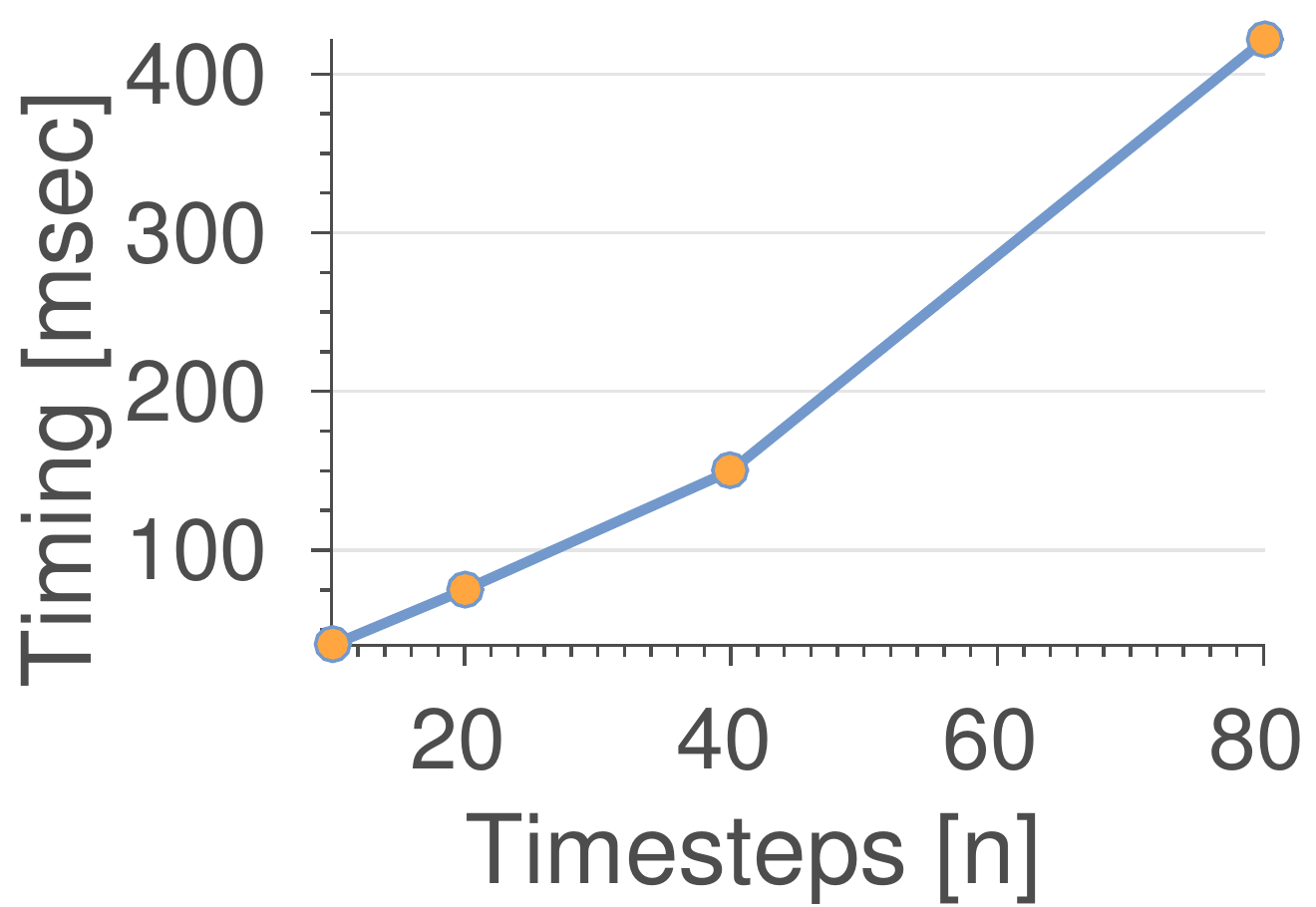}
    \par\vspace{0pt}
  \end{minipage}
 \caption{\small Average time required to construct and solve the dynamics optimization problem given a fixed set of footsteps.}
 \label{table_timing_statistics}
 \vspace{-0.4cm}
\end{figure}

Table \ref{table_timing_statistics} reports average time required in our formulation to build and solve the dynamics optimization problem (convex QCQP) given a fixed set of contacts for different horizon lengths. We used the "Climbing an uneven terrain using hands" tasks
because it is the task with the highest computational complexity.
As can be seen in the table and the graph, thanks to the problem sparsity, the complexity does not deteriorate too quickly
and for the operating conditions that we use, complexity increases close to linearly.
If we use a dynamics granularity of 200ms (40 timesteps), it takes 142ms to find a 8 second long plan involving 4 contact changes.
If we increase the dynamics granularity to 100ms (80 timesteps), it takes us around 421ms to solve the problem and we could solve it 2 times in a second. Another way to interpret this result is as follows: if we were to plan for a time horizon of 2 sec (maybe 2 footsteps), using a discretization step of 100ms (20 timesteps) we could solve the problem in around 85ms using the same granularity of 100ms, we could plan for an horizon of half a minute (320 timesteps) in 3 sec. 

\subsection{Tightness of the solution}
As mentioned earlier, using the cost as defined in section \ref{subsec:cost_function}, 
all our numerical experiments found that the constraint relaxation is tight:
the gap between upper, lower bounds and their corresponding true values is zero.
It means that in these cases we found solutions that are also an optimal solution to the original (non-convex) problem.

This result is potentially very interesting because if the relaxation was always tight, it would mean that
it is always possible to find the global solution of the non-convex momentum optimization problem by solving a convex QCQP.
In the optimization literature, there are a few results about strong duality in non-convex quadratic optimization such as \cite{Nonconvex01}. There is for example one strong result in the case when the Hessian of the objective has a null-space
where it is shown for a simple case with a quadratic objective and a quadratic equality constraint, that because of the presence of this nullspace, the relaxation is tight (zero gap) and optimal.
Since our objective function does not minimize for the angular momentum, there is also a nullspace, it might be possible
that similar arguments could be used. However, a formal proof remains for future work.

\section{CONCLUSION} \label{sec:conclusion}
We have proposed a convex relaxation of the momentum dynamics for leg robots that can be used to efficiently plan contact forces, CoM motion and momentum trajectories. Moreover, we have proposed an extension of a contact planning algorithm
including our convex model to find dynamically consistent contact sequences. 
Computation times of the algorithm are small enough to be used in a receding horizon fashion, moreover, 
numerical experiments suggest that the convex relaxation is tight. While a formal proof is missing,
our result suggests that momentum and contact forces optimization, a non-convex problem, 
could be solved exactly using our convex relaxation, therefore removing the need for further model simplification.
\bibliographystyle{plain}
\bibliography{references}

\begin{thebibliography}{10}

\bibitem{DBLP:conf/iros/AudrenVKEKY14}
H.~Audren, J.~Vaillant, A.~Kheddar, A.~Escande, K.~Kaneko, and Yoshida. E.
\newblock Model preview control in multi-contact motion-application to a
  humanoid robot.
\newblock In {\em IROS}, pages 4030--4035, 2014.

\bibitem{ConvexOptimization}
S.~Boyd and L.~Vandenberghe.
\newblock {\em Convex Optimization}.
\newblock Cambridge University Press, New York, 2009.

\bibitem{NonsmoothMechanics}
B.~Brogliato.
\newblock {\em Nonsmooth Mechanics - Models, Dynamics and Control}.
\newblock Communications and Control Engineering. Springer-Verlag London, 2
  edition, 1999.

\bibitem{LinearSystemsTheory}
F.~Callier and C.~Desoer.
\newblock {\em Linear Systems Theory}.
\newblock Springer, New York, 1994.

\bibitem{DBLP:conf/icra/CaronPN15}
S.~Caron, Q.C. Pham, and Y.~Nakamura.
\newblock Stability of surface contacts for humanoid robots: Closed-form
  formulae of the contact wrench cone for rectangular support areas.
\newblock In {\em ICRA}, pages 5107--5112.

\bibitem{JustinMomentumOptimization}
J.~Carpentier, S.~Tonneau, M.~Naveau, O.~Stasse, and N.~Mansard.
\newblock A versatile and efficient pattern generator for generalized legged
  locomotion.
\newblock In {\em ICRA}, 2016.

\bibitem{DBLP:conf/humanoids/DaiVT14}
H.~Dai, A.~Valenzuela, and R.~Tedrake.
\newblock Whole-body motion planning with centroidal dynamics and full
  kinematics.
\newblock In {\em Humanoids}, 2014.

\bibitem{DBLP:conf/humanoids/DeitsT14}
R.~Deits and R.~Tedrake.
\newblock Footstep planning on uneven terrain with mixed-integer convex
  optimization.
\newblock In {\em Humanoids, Madrid-Spain}, pages 279--286, 2014.

\bibitem{conf/icra/DimitrovPW11}
D.~Dimitrov, A.~Paolillo, and P.B. Wieber.
\newblock Walking motion generation with online foot position adaptation based
  on ℓ1- and ℓ℞-norm penalty formulations.
\newblock In {\em ICRA}, pages 3523--3529, 2011.

\bibitem{journals/trob/EnglsbergerOA15}
J.~Englsberger, C.~Ott, and A.~Albu-Schäffer.
\newblock Three-dimensional bipedal walking control based on divergent
  component of motion.
\newblock {\em IEEE Trans. Robotics}, 31(2):355--368, 2015.

\bibitem{AlexAuroPaper}
A.~Herzog, N.~Rotella, S.~Mason, F.~Grimminger, S.~Schaal, and L.~Righetti.
\newblock Momentum control with hierarchical inverse dynamics on a
  torque-controlled humanoid.
\newblock {\em Auton. Robots}, 40(3):473--491, 2016.

\bibitem{AlexHumanoidsPaper}
A~Herzog, N~Rotella, S~Schaal, and L~Righetti.
\newblock Trajectory generation for multi-contact momentum-control.
\newblock In {\em Humanoids}, 2015.

\bibitem{Herzog-2016b}
A.~Herzog, S.~Schaal, and L.~Righetti.
\newblock Structured contact force optimization for kino-dynamic motion
  generation.
\newblock 2016.

\bibitem{ibanez-IROS2014}
A.~Ibanez, P.~Bidaud, and V.~Padois.
\newblock Emergence of humanoid walking behaviors from mixed-integer model
  predictive control.
\newblock In {\em IROS}, 2014.

\bibitem{conf/icra/KajitaKKFHYH03}
F.~Kajita, S.and~Kanehiro, K.~Kaneko, K.~Fujiwara, K.~Harada, K.~Yokoi, and
  H.~Hirukawa.
\newblock Biped walking pattern generation by using preview control of
  zero-moment point.
\newblock In {\em ICRA}, pages 1620--1626. IEEE, 2003.

\bibitem{journals/tog/MordatchTP12}
I.~Mordatch, E.~Todorov, and Z.~Popovic.
\newblock Discovery of complex behaviors through contact-invariant
  optimization.
\newblock {\em ACM Trans. Graph.}, 31(4):43:1--43:8, 2012.

\bibitem{InteriorPointPTM}
A.~Nemirovski.
\newblock {\em Interior Point Polynomial Time Methods in Convex Programming}.
\newblock 2004.

\bibitem{DBLP:conf/iros/Wieber08}
Wieber. P.B.
\newblock Viability and predictive control for safe locomotion.
\newblock In {\em IROS}, pages 1103--1108, 2008.

\bibitem{DBLP:conf/wafr/PosaT12}
M.~Posa and R.~Tedrake.
\newblock Direct trajectory optimization of rigid body dynamical systems
  through contact.
\newblock In {\em WAFR}, 2012.

\bibitem{Nonconvex01}
M.~Salahi.
\newblock Convex optimization approach to a single quadratically constrained
  quadratic minimization problem.
\newblock {\em Central European J of Operations Res}, 18(2):181--187, 2010.

\bibitem{DBLP:journals/tsmc/SardainB04}
P.~Sardain and G.~Bessonnet.
\newblock Forces acting on a biped robot. center of pressure-zero moment point.
\newblock {\em {IEEE} Trans. Systems, Man, and Cybernetics, Part {A}},
  34(5):630--637, 2004.

\bibitem{DBLP:conf/humanoids/SherikovDW14}
A.~Sherikov, D.~Dimitrov, and P.B. Wieber.
\newblock Whole body motion controller with long-term balance constraints.
\newblock In {\em Humanoids}, pages 444--450, 2014.

\bibitem{DBLP:conf/iros/TassaET12}
Y.~Tassa, T.~Erez, and E.~Todorov.
\newblock Synthesis and stabilization of complex behaviors through online
  trajectory optimization.
\newblock In {\em IROS}, 2012.

\bibitem{WieberNonholonomy}
P.-B. Wieber.
\newblock Holonomy and nonholonomy in the dynamics of articulated motion.
\newblock {\em Fast Motions in Biomechanics and Robotics}, pages 411--425,
  2006.

\bibitem{conf/humanoids/Wieber06}
P.B. Wieber.
\newblock Trajectory free linear model predictive control for stable walking in
  the presence of strong perturbations.
\newblock In {\em Humanoids}, pages 137--142, 2006.

\end{thebibliography}
\end{document}